\documentclass[10pt,journal,compsoc]{IEEEtran}

\ifCLASSOPTIONcompsoc
  \usepackage[nocompress]{cite}
\else
  \usepackage{cite}
\fi

\usepackage{times}
\usepackage{epsfig}
\usepackage{graphicx}
\usepackage{amsmath}
\usepackage{amssymb}
\usepackage{dsfont}
\usepackage{bm}
\usepackage{marvosym}
\usepackage{ulem}

\usepackage[pagebackref=true,breaklinks=true,colorlinks,bookmarks=false]{hyperref}

\begin{document}
%
\title{Robust Point Cloud Registration Framework Based on Deep Graph Matching}

\author{Kexue Fu\textsuperscript{*},
        Jiazheng Luo\textsuperscript{*},
        Xiaoyuan Luo,
        Shaolei Liu,
        Chenxi Zhang\textsuperscript{\Letter},
        Manning Wang\textsuperscript{\Letter},
\IEEEcompsocitemizethanks{\IEEEcompsocthanksitem Kexue Fu, Jiazheng Luo, Xiaoyuan Luo, Shaolei Liu, Chenxi Zhang and Manning Wang are with the Digital Medical Research Center, School of Basic Medical Science, Fudan University, Shanghai 200032, China, and also with the Shanghai Key Laboratory of Medical Image Computing and Computer Assisted Intervention, Shanghai, China, 200032.\protect\\
* These authors have contributed equally to this work.\protect\\
\Letter Corresponding author.\protect\\
E-mail: \{fukexue, jzluo20, xyluo19, liushaolei, chenxizhang, mnwang\}@fudan.edu.cn \protect\\
}
\thanks{Manuscript received xxxxx xxxx, xxxx; revised xxxx xx, xxxx.}}

\markboth{Journal of \LaTeX\ Class Files,~Vol.~xx, No.~x, xxxxx~xxxx}%
{Shell \MakeLowercase{\textit{et al.}}: Bare Demo of IEEEtran.cls for Computer Society Journals}

\IEEEtitleabstractindextext{%
\begin{abstract}
 3D point cloud registration is a fundamental problem in computer vision and robotics. Recently, learning-based point cloud registration methods have made great progress. However, these methods are sensitive to outliers, which lead to more incorrect correspondences. In this paper, we propose a novel deep graph matching-based framework for point cloud registration. Specifically, we first transform point clouds into graphs and extract deep features for each point. Then, we develop a module based on deep graph matching to calculate a soft correspondence matrix. By using graph matching, not only the local geometry of each point but also its structure and topology in a larger range are considered in establishing correspondences, so that more correct correspondences are found. We train the network with a loss directly defined on the correspondences, and in the test stage the soft correspondences are transformed into hard one-to-one correspondences so that registration can be performed by a correspondence-based solver. Furthermore, we introduce a transformer-based method to generate edges for graph construction, which further improves the quality of the correspondences. Extensive experiments on 
object-level and scene-level benchmark datasets show that the proposed method achieves state-of-the-art performance. The code is available at: \href{https://github.com/fukexue/RGM}{https://github.com/fukexue/RGM}.
\end{abstract}

\begin{IEEEkeywords}
Point Cloud Registration, Graph Matching, Correspondence.
\end{IEEEkeywords}}

\maketitle

\IEEEdisplaynontitleabstractindextext

\IEEEraisesectionheading{\section{Introduction}\label{sec:introduction}}

\IEEEPARstart{R}{igid} point cloud registration is a task that finds a rigid transformation to align two point clouds, and it has long been a fundamental task in computer vision and robotics, with many important applications, such as autopilot \cite{ref1,ref2,ref56}, surgical navigation \cite{ref3} and SLAM \cite{ref4,ref5}. There are two interlocked subproblems in point cloud registration: finding the transformation to align the two point clouds and finding the correspondences between the points \cite{ref6}. Although when the solution to one subproblem is known, the other subproblem can be easily solved, it is difficult to solve both subproblems together. Point cloud registration becomes even harder when there are outliers, which are the points with no correspondences in the other point cloud. Outliers may come from the imperfectness of the sensors used to collect the point clouds or situations in which the two point clouds to be registered are not fully overlapped.

\begin{figure}[t]
   \begin{center}
      \includegraphics[width=1\linewidth]{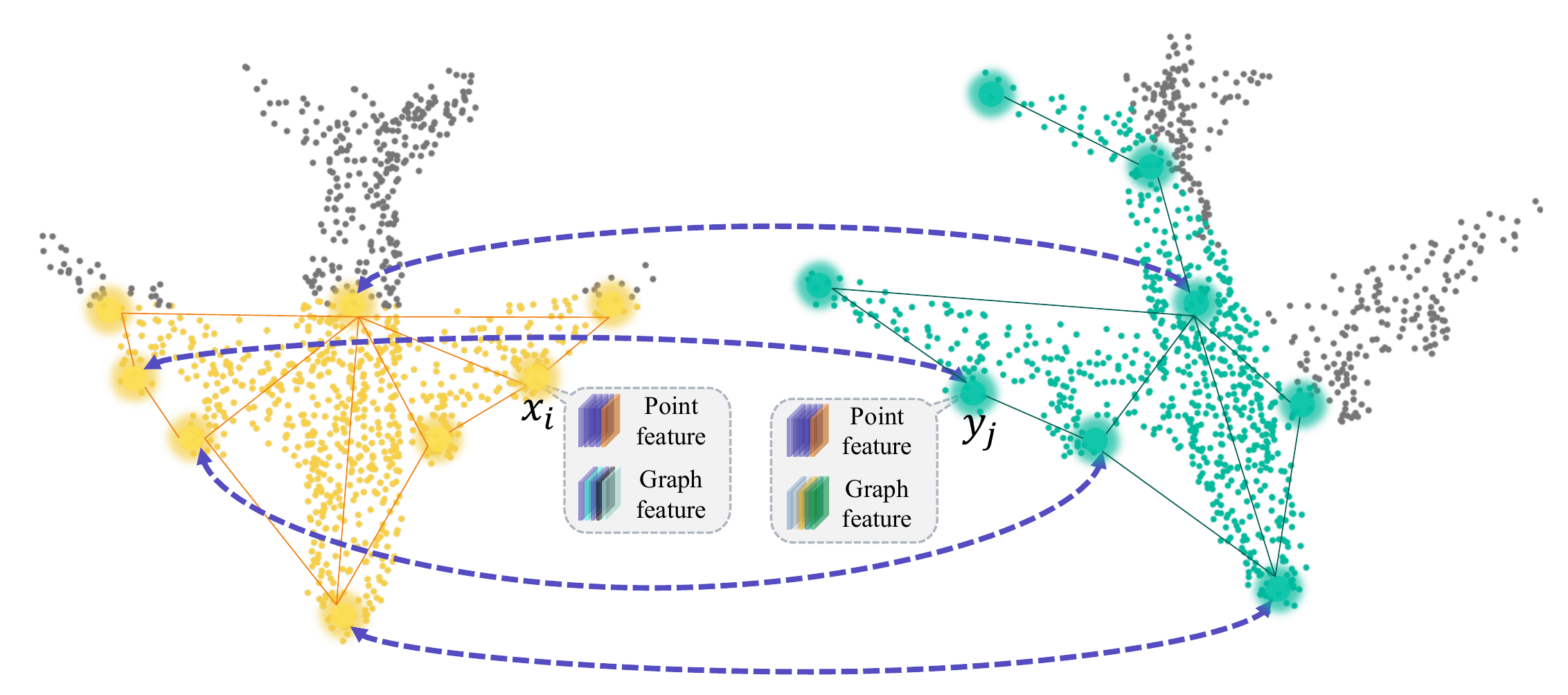}
   \end{center}
      \caption{The idea of point cloud registration based on graph matching. Dashed lines represent correspondences. Point features and graph features are the features extracted directly through points and the features extracted based on graphs, respectively. The two points $x_i$ and $y_j$ have similar point features because they have similar local geometries, but they have different graph features because the graph topologies around them are different, so they are not mismatched when graph-based matching is used.}
   \label{fig1}
\end{figure}

Iterative closest point (ICP) \cite{ref7,ref8} is arguably the most widely used method for rigid point cloud registration, which starts from an initial transformation and alternately updates the correspondences and transformation. One major limitation of ICP is that it can only converge to a local optimum near the initialization, and its convergence basin is fairly small, especially when there are noise and outliers.  A series of global registration methods based on branch-and-bound (BnB) \cite{ref9,ref10,ref11} have been proposed to alleviate the need for initialization by obtaining the global optimal solution, but the time-consuming BnB limits their practical applications. Another method for mitigating the need for initialization is through keypoint extraction and matching \cite{ref12,ref13}. However, these methods \cite{ref12,ref13} are sensitive to outliers and repetitive geometry \cite{ref14}. 

The recent registration studies have been dominated by learning-based methods, most of which establish an end-to-end trainable network. Some of them learn the global features of each point cloud and regress the transformation directly from those global features \cite{pointnetlk1, pointnetlk2}. Other methods integrate neural networks that predict soft correspondence and a differentiable singular value decomposition (SVD) algorithm to solve transformation, such as DCP \cite{ref15}, RPM-Net \cite{ref16} and IDAM \cite{ref17}, and they do not need transformation initialization. Although these end-to-end methods have made great progress in object-level registration, they do not work well in scene-level registration \cite{ref58}. Generally, scene-level registration \cite{ref58,ref60,ref61,ref65,ref62} begins with a neural network to predict hard correspondences and then solves the transformation using RANdom SAmple Consensus (RANSAC) \cite{ref64}. These methods explore deep features to establish correspondences but the discrimination ability of the features extracted from point clouds is limited, as shown in Figure~\ref{fig1}, which leads to a large proportion of outliers in the putative correspondences and consequently devastates the registration accuracy.

In this paper, we propose a robust point cloud registration framework that utilizes deep graph matching to better handle outliers, and we denote it as RGM (Registration by Graph Matching). By constructing graphs from point clouds to be registered and capturing  the high-order structure of the graphs, RGM can find robust and accurate point-to-point correspondences to better solve the point cloud registration problem. To the best of our knowledge, this is the first time that deep graph matching has been applied to point cloud registration. RGM contains an end-to-end deep neural network, the first part of which is a feature extractor that extracts deep local features for each point by using its neighboring points. Instead of matching these local point features directly, we construct a graph for each of the two point clouds and embed both the graph nodes (local features for each point) and graph structure (second-order or high-order structure) into node feature space \cite{ref18}. Then, we introduce a module consisting of an affinity layer, instance normalization and Sinkhorn to predict soft correspondences from the node features of the two graphs, and we denote it as AIS module. By using graph matching in the AIS module, not only the local geometry of each node but also its structure and topology in a larger range are considered in establishing correspondences so that more correct correspondences are found. In training, the focal loss between the predicted soft correspondences and the ground-truth correspondences are adopted, which directly promotes the network to learn better point-to-point correspondences. In testing, we use the linear assignment problem (LAP) solver \cite{ref52} based on the Hungarian algorithm \cite{ref40} to transform soft correspondences into one-to-one hard correspondences, and then a correspondence-based estimator is employed to calculate the transformation from the hard correspondences. Similar to existing methods such as RPM-Net and ICP, we iteratively optimize the registration results. 

Our main contributions are as follows:
\begin{itemize}
   \item We propose using deep graph matching to solve the point cloud registration problem for the first time. Instead of only using the features of each point, graph matching can leverage the features of other nodes and the structural information of graphs when establishing correspondences so that it can better address the problem of outliers.
   \item We propose using a transformer to generate soft graph edges at the beginning of each iteration. Utilizing the attention and co-attention mechanism in the transformer, more and more meaningful edges are built in the overlapping regions of the two point clouds. As a result, better correspondences can be established for the overlapping parts in registering partial-to-partial point clouds.
   \item We introduce the AIS module to establish reliable correspondences between nodes of two given graphs. The AIS module calculates an affinity matrix between the nodes of the two graphs based on the embedded features, and by analyzing the affinity matrix globally and utilizing the Sinkhorn algorithm, it can effectively reduce the proportion of incorrect correspondences.
   \item Our method achieves state-of-the-art performance on both object-level and scene-level benchmark datasets.
\end{itemize}

A preliminary version of this paper was published in conference \cite{ourcvpr}. This paper significantly improves the previous work to make it applicable in more difficult scenarios and have better performance. Concretely, (i) We change the Binary Cross-Entropy Loss to Focal Loss \cite{lin2017focal} to make the network pay more attention to the hardest correspondences. (ii) Instead of only using SVD as the transformation estimator, we extend it to use a more robust correspondence-based estimator, RANSAC. (iii) We add a new local feature extractor for scene-level data. (iv) Many new experiments have been conducted on the improved methods, including cross-dataset generalization, registration under full-range rotation, and scene-level registration experiments on 3DMatch and 3DLoMatch. Experimental results show that our method significantly outperforms the state-of-the-art methods.

\section{Related Work}
\subsection{Traditional Registration Method}
A large proportion of traditional methods need an initial transformation and find a locally optimal solution near the initialization, in which ICP \cite{ref7,ref8} is an early and representative method. ICP starts with an initial transformation and iteratively alternates between solving two trivial subproblems: finding the closest points as correspondences under current transformation and computing optimal transformation by SVD from found correspondences. Many variants have been proposed to improve ICP \cite{ref19,ref20,ref21}. Nevertheless, ICP and its variants can only converge to a local optimum, and their success heavily relies on a good initialization. To improve the robustness to noise and outliers and enlarge the convergence basin, some methods transform point clouds into probability distributions and reformulate point cloud registration as matching two probability distributions, such as GMM \cite{ref22} and HGMR \cite{ref23}. These methods do not need to alternately solve correspondences and transformation, but their objective functions are nonconvex, so they still need a good initialization to avoid converging to a bad local optimum. Recently, a series of globally optimal methods based on BnB have been proposed, such as Go-ICP \cite{ref9}, GOGMA \cite{ref10}, GOSMA \cite{ref11}, and GoTS \cite{ref24}, but they are very slow and only practical in some limited scenarios. Another line of work avoids transformation initialization by establishing correspondences. They usually first extract keypoints from the original point clouds and construct feature descriptors for them and then establish potential correspondences through feature matching \cite{ref12,ref13}. After that, RANSAC-like algorithms can be used to find the correct correspondences for registration. Different from RANSAC-like methods, FGR \cite{ref25} optimizes a correspondence-based objective function by a graduated nonconvex strategy and achieves state-of-the-art performance in correspondence-based point cloud registration. However, correspondence-based methods are sensitive to duplicate structures and partial-to-partial point clouds because a large proportion of the potential correspondences will be incorrect in these scenarios. Specifically, the lack of good initialization, a large proportion of outliers and time constraints are still big challenges for traditional point cloud registration methods.

\subsection{Direct Deep Registration Method}
The developments of deep learning on point clouds processing allow researchers to make good use of existing techniques, such as PointNet \cite{ref26}, and DGCNN \cite{ref27}, to extract point cloud features for downstream tasks. These studies have stimulated the interest of using deep learning in point cloud registration. One of the earliest works is PointNetLK \cite{ref29}, which calculates global feature descriptors of the two point clouds through PointNet and iteratively uses the IC-LK algorithm \cite{ref30,ref31} to minimize the distance between the two global feature descriptors to achieve registration. PCRNet \cite{ref14} replaces the IC-LK algorithm in PointNetLK with a deep neural network. DCP \cite{ref15} utilizes transformer \cite{ref42,ref43} to compute soft correspondences between two point clouds and utilizes a differentiable SVD algorithm to calculate the transformation. Although these methods have the advantages of being fast and some of them do not need transformation initialization, they cannot effectively handle partial-to-partial point cloud registration. PRNet \cite{ref34} proposes a keypoint detector and uses the keypoint-to-keypoint correspondences in a self-supervised way to solve the partial-to-partial point cloud registration. DeepGMR \cite{ref35} extracts pose-invariant correspondences between raw point clouds and Gaussian mixture model (GMM) parameters, and then recovers the transformation from the matched Gaussian mixture models. IDAM \cite{ref17} integrates the iterative distance-aware similarity convolution module into the matching process, which can overcome the shortcomings of using inner products to obtain pointwise similarity. RPM-Net \cite{ref16} proposes a network to predict optimal annealing parameters and uses annealing and Sinkhorn \cite{ref36} to obtain soft correspondences from local features. Soft correspondences can increase robustness, but they lead to the decrease of registration accuracy, which is shown in our experiments on clean point clouds. Although these methods can handle partial-to-partial point cloud registration to some extent, there is still room for improvement in their accuracy and robustness. The difference between our method and the existing learning-based methods is that we construct graphs from the original point clouds and merge structural information of the graphs into node features so that the nodes can be better matched.

\subsection{Feature-based Deep Registration Method}
Most of the direct point cloud registration methods used in object-level registration do not work well in scene-level registration. For scene-level registration, it has become popular to build correspondences based on the learned local feature descriptors of point clouds first, and then compute the transformation using the robust estimator RANSAC \cite{ref64}. These methods focus on how to obtain more efficient and robust local features. For example, 3DMatch \cite{ref65} converts point cloud patches into volumetric voxel grids and uses a 3D convolution network to extract local features. The benchmark dataset presented in 3DMatch  \cite{ref65} is popular in evaluating scene-level registration algorithms. PPFNet \cite{ref66} presented a 3D local patch descriptor with an augmented set of simple geometric relationships: points, normals and point pair features (PPF). These two methods \cite{ref65,ref66} rely on the supervision of correspondences between patches. Later methods utilize self-supervised or unsupervised learning for feature extraction. For example, PPF-FoldNet \cite{ref68} and CapsuleNet \cite{ref69} train an autoencoder to reconstruct the original input point clouds without supervision and use the trained encoder for feature extraction.
However, features are extracted only for predefined patches, so the perception field of these methods \cite{ref65, ref66, ref68, ref69} is greatly limited. Therefore, computing the feature descriptors for each point is a better option.
The key issue is how to efficiently extract point features for large scene-level point clouds. FCGF \cite{ref60} uses 3D sparse convolution to process the complete point cloud to generate dense feature predictions, which are more efficient than patch-based methods. Similar to FCGF, D3Feat \cite{ref61} uses a fully convolution neural network KPConv \cite{thomas2019kpconv} to compute dense features, and it jointly predicts dense features and point saliency score. Predator \cite{ref58} utilizes cross attention mechanism to obtain more robust feature descriptors. StickyPillars \cite{fischer2021stickypillars} use a graph neural network and the Sinkhorn algorithm to solve outdoor scene registration. Different from PREDATOR and StickyPillars, we designed an edge generator to update the features of points in overlapping areas using the features of overlapping areas, while features of points in non-overlapping areas are updated only by themselves. This prevents the features of overlapping areas from being disturbed by features of non-overlapping areas. 

\subsection{Learning for Graph Matching}
Graph matching has been widely studied in computer vision and pattern recognition \cite{ref47,ref50,ref51,ref57,sarlin2020superglue}. Recently, learning-based graph matching has attracted considerable research interest \cite{ref37,ref38,ref18}, but, to the best of our knowledge, there is no research on using learning-based graph matching to solve the point cloud registration problem.

\begin{figure*}[t]
   \centering
   \includegraphics[width=1\linewidth]{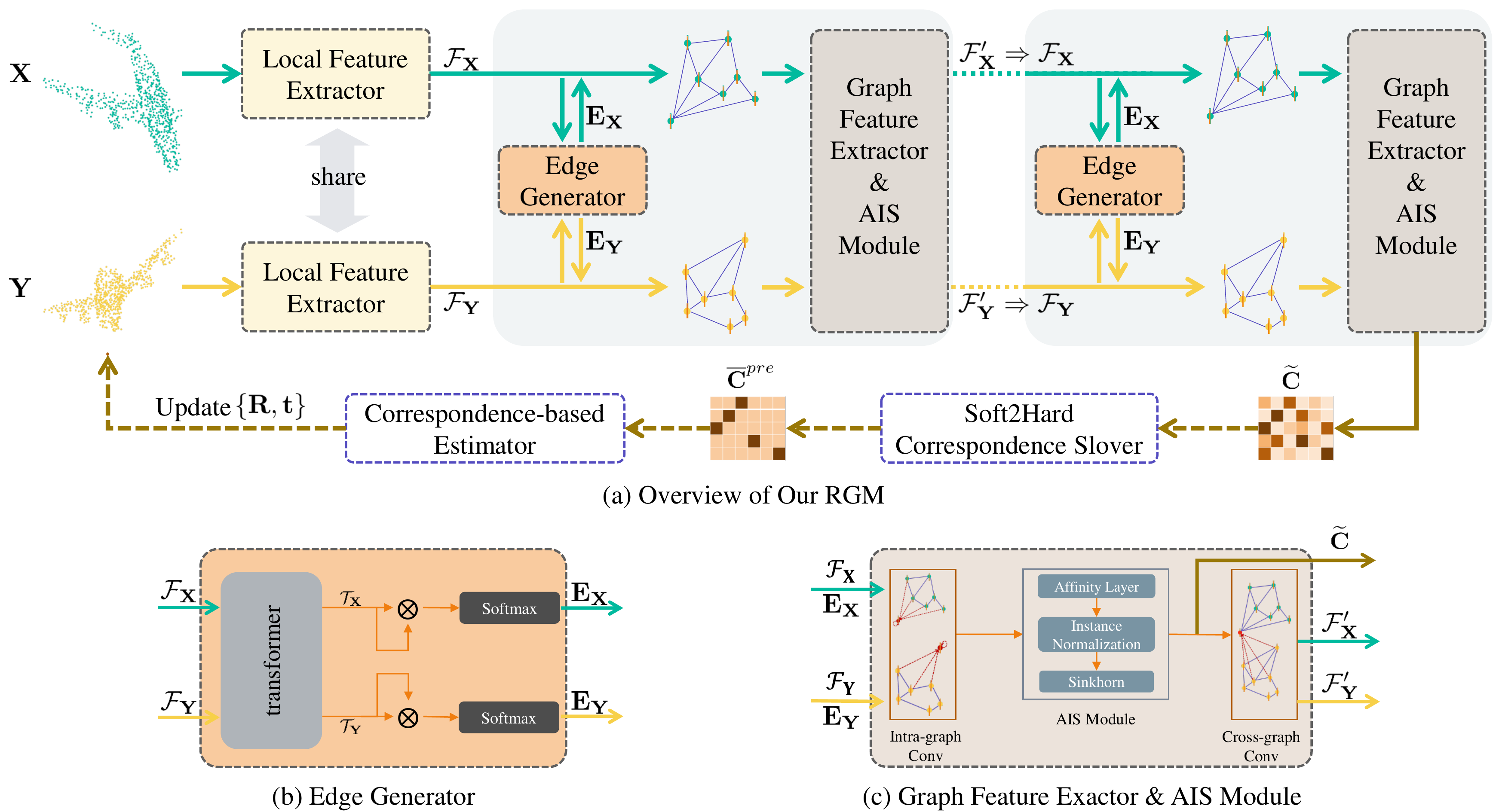}
   \caption{The pipeline of the proposed 3D rigid point cloud registration framework, RGM, where $\bigoplus$ represents concatenate features and $\bigotimes$ represents matrix multiplication. The solid lines are the data flow of both training and testing, and the dashed lines are the data flow that exists only in testing.}
   \label{fig2}%
\end{figure*}

\section{Problem Formulation}
3D rigid point cloud registration refers to estimating a rigid transformation $\mathbf{T}\in\mathbf{SE}(3)$ with parameters $\mathbf{R}\in\mathbf{SO}(3)$, $\mathbf{t}\in\mathbb{R}^3$ to align a source point cloud $\mathbf{X}=\left\{x_i\in\mathbb{R}^3|i=1,\cdots,N\right\}$ and a target point cloud $\mathbf{Y}=\left\{y_j\in\mathbb{R}^3|j=1,\cdots,M\right\}$. $N$ and $M$ represent the number of points in $\mathbf{X}$ and $\mathbf{Y}$, respectively. The correspondences between points in $\mathbf{X}$ and $\mathbf{Y}$ are represented by matrix $\mathbf{C}=\left\{0,1\right\}^{N\times M}$. If $x_i$ and $y_j$ are a pair of corresponding points, $\mathbf{C}_{i,j}$ is 1; otherwise, it is 0. We first consider the simple case where there are strict one-to-one correspondences between points in $\mathbf{X}$ and $\mathbf{Y}$, in which, $N=M$. The rigid point cloud registration problem can be formulated as minimizing the following objective function:
\begin{equation}
   \boldsymbol{e}(\mathbf{C}, \mathbf{R}, \mathbf{t})=\sum_{i}^{N} \sum_{j}^{M} \mathbf{C}_{i, j}\left\|\mathbf{R} x_{i}+\mathbf{t}-y_{j}\right\|_{2}^{2},
\end{equation}
$\text {subject to } \sum_{j}^{M} \mathbf{C}_{i, j}=1, \forall i,\ \sum_{i}^{N} \mathbf{C}_{i, j}=1, \forall j,\ \mathbf{C}_{i, j} \in\{0, 1\}^{N \times M}, \forall i, j$. In the more difficult case where there are no one-to-one correspondences, the equality constraints no longer hold, and they become inequality constraints. We can introduce slack variables in $\mathbf{C}$ as in \cite{ref16} to convert inequality constraints back into equality constraints. The row constraints are converted as follows, and the column constraints are similarly converted:
\begin{equation}
   \sum_{j}^{M} \mathbf{C}_{i, j} \leq 1, \forall i \rightarrow \sum_{j}^{M+1} \mathbf{C}_{i, j}=1, \forall i \leq N.
\end{equation}
Please note that $\mathbf{C}$ becomes a $(N+1)\times(M+1)$ matrix after introducing one row and one column slack variables, and the sums of the added row and column are not restricted to be one.

In this paper, we use an end-to-end neural network to predict $\mathbf{C}$. Once we know the correspondences, the rigid transformation can be obtained by a correspondence-based solver.

\section{RGM}
Figure~\ref{fig2} (a) shows the overall pipeline of RGM. RGM consists of five components: local feature extractor, edge generator, graph feature extractor $\&$ AIS module, Soft2Hard Correspondence Solver and Correspondence-based Estimator. During training, we use the shared local feature extractor to extract local features for each point in $\mathbf{X}$ and $\mathbf{Y}$, and take these local features as the node features $\mathcal{F}$ of the initial graph. Next, the edge generator generates edges and builds the source graph and target graph, and the graphs are inputted into the graph feature extractor, which processes the two graphs and outputs new node features $\mathcal{F}^\prime$ and uses them to update $\mathcal{F}$. The AIS module predicts the soft correspondence matrix $\widetilde{\mathbf{C}}$ between nodes of the two graphs. By using blocks composed of three modules, the edge generator, graph feature extractor and AIS module, with the same structure but different weights $L$ times, we can obtain node features $\mathcal{F}$ with better discrimination capability and a more accurate soft correspondence matrix $\widetilde{\mathbf{C}}$. Finally, we adopt the focal loss between $\widetilde{\mathbf{C}}$ and the ground truth correspondences to train the network. During test, two point clouds are first inputted into the network to obtain the soft correspondence matrix $\widetilde{\mathbf{C}}$. Then, the soft correspondences are converted to hard correspondences using the Soft2Hard Correspondence Solver, and the transformation is solved by a Correspondence-based Estimator. Similar to ICP, we also update the transformation iteratively during test time. The details of each component are explained in the following subsections.

\subsection{Local Feature Extractor}
\label{sec:feature_extractor} 
To establish the correspondence matrix between two point clouds, it is necessary to embed the source point cloud $\mathbf{X}$ and the target point cloud $\mathbf{Y}$ into a common feature space. We choose different backbones to extract high-dimensional features to better fit different kinds of datasets. For object-level datasets, we first define the local feature descriptor $\mathcal{P}_{x_i}$ of $x_i$ as:
$\mathcal{P}_{x_{i}}=\left\{\left(x_{i}, x_{n}\right) \mid \forall x_{n} \in \mathcal{K}_{i}\right\},$where $\mathcal{K}_i$ represents the $K$-nearest neighboring points of $x_i$. Then low-dimensional local feature descriptors are mapped to high-dimensional local feature spaces through nonlinear functions $f_\theta$: $\mathbb{R}^{K\times6}\rightarrow\mathbb{R}^V$, where $V$ is the dimensionality of the final high-dimensional local feature. The implementation of $f_\theta$ can be found in Section 1 of the supplementary materials, where $\theta$ represents the parameter of the nonlinear function, which consists of shared multilayer perceptron (MLP), maxpooling and concatenation. We use the high-dimensional local features as the node features $\mathcal{F}$ of the initial graph. The node feature $\mathcal{F}_{x_i}$ of $x_i$ can be expressed as follows:
$\mathcal{F}_{x_{i}}=f_{\theta}\left(\mathcal{P}_{x_{i}}\right), \mathcal{F}_{x_{i}} \in \mathbb{R}^V.$ 

Unlike object-level point clouds, scene-level point clouds contain a great number of points and have more complex geometric structure. To improve computational efficiency and feature representation capabilities on scene-level point clouds, we build the local feature extractor $f_\theta$ based on KPFCN \cite{thomas2019kpconv} to extract high-dimensional local features. The local feature extractor takes the source point cloud $\mathbf{X}$ and the target point cloud $\mathbf{Y}$ as inputs, and outputs the coordinates and local features of the new source $\mathbf{X}$ and the new target point cloud $\mathbf{Y}$, which are obtained by layer-by-layer grid-based subsampling. KPFCN uses kernel point convolution to extract the local features of each points, which can better learn the geometric information of neighbor points. In addition, it uses multi-layer grid-based subsampling to reduce the number of points and expand the feature receptive field. For more details about the implementation of KPFCN, we recommend reading the original paper \cite{thomas2019kpconv}. The above process can be formulated as $(\mathbf{X}, \mathcal{F}_{x_{i}})=f_\theta(\mathbf{X})$. In order to balance computation cost and registration accuracy, we constructed the local feature extractor by removing the decoder units from the 2nd to the last un-pooling layer from the default UNet-like KPFCN structure.

Inspired by the idea of the Siamese network \cite{ref41}, the two point clouds share the same local feature extractor. When the two point clouds become closer, the local features also become similar, so this structure is suitable for iterative registration.

If only the local features are used to predict the correspondences between point clouds, it is easy to obtain incorrect correspondences, especially when there are outliers. The reason is that the local features do not contain the structural information of the point cloud on a larger scale (self-correlation) and the association between the two point clouds (cross-correlation). Inspired by Wang’s research on deep graph matching \cite{ref18}, we construct graphs from point clouds and use deep graph matching to establish better correspondences. Section \ref{secEdge} describes how to build graphs from point clouds, and Section \ref{secGraph} introduces how to predict the correspondences by using deep graph matching and the AIS module.


\subsection{Transformer based Edge Generator}\label{secEdge}
The graphs built from $\mathbf{X}$ and $\mathbf{Y}$ are denoted as source graph $\mathcal{G}_s=\left\{\mathbf{X},\mathbf{E}_\mathbf{X}\right\}$ and target graph $\mathcal{G}_t=\left\{\mathbf{Y},\mathbf{E}_\mathbf{Y}\right\}$, respectively. The graph nodes are the original points, and the graph edges are represented by the adjacency matrix $\mathbf{E}$. The node features of $\mathcal{G}_s$ and $\mathcal{G}_t$ are denoted by $\mathcal{F}_{x_i}$ and $\mathcal{F}_{y_j}$, respectively. There are trivial methods to generate the edges, such as full connection, nearest neighbor connection and Delaunay triangulation, but the features of graphs cannot be effectively aggregated, resulting in poor correspondences, as shown in Figure~\ref{fig5} (d). Inspired by the success of BERT \cite{ref42} in NLP, we introduce a transformer \cite{ref43} module to dynamically learn the soft edges of any two nodes within a point cloud. The transformer-based edge generator is illustrated in Figure~\ref{fig2} (b). Detail structure of the edge generator can be found in Section 1.3 of the supplementary materials. The transformer consists of several stacked encoder-decoder layers. The encoder uses a self-attention layer and shared MLP to encode node features, and the decoder associates and encodes features based on the co-attention mechanism. The transformer takes node features $\mathcal{F}_\mathbf{X},\mathcal{F}_\mathbf{Y}$ as input and encodes them into embedding features $\mathcal{T}_\mathbf{X},\mathcal{T}_\mathbf{Y}$. Soft edge adjacency matrices are obtained by applying a softmax function on the inner product of the embedding features as follows:
\begin{equation}
   \mathcal{T}_{\mathbf{X}}, \mathcal{T}_{\mathbf{Y}}=f_{\text {transformer}}(\mathcal{F}_{\mathbf{X}}, \mathcal{F}_{\mathbf{Y}}),
\end{equation}
\begin{equation}
   \mathbf{E}_{\mathbf{X}}=\operatorname{softmax}(\langle\left(\mathcal{T}_{\mathbf{X}}\right)^{T}, \mathcal{T}_{\mathbf{X}}\rangle),
\end{equation}
\begin{equation}
   \mathbf{E}_{\mathbf{Y}}=\operatorname{softmax}(\langle\left(\mathcal{T}_{\mathbf{Y}}\right)^{T}, \mathcal{T}_{\mathbf{Y}}\rangle).
\end{equation}
\subsection{Graph Feature Extractor and AIS Module} \label{secGraph}
This part is shown in Figure~\ref{fig2} (c), which consists of three consecutive steps as follows:
First, we use intra-graph conv to explore the self-correlation of node features, where features are aggregated from nodes along edges within each graph. The message passing scheme between nodes is the same as PCA-GM \cite{ref18}. A node self-correlation feature $\mathcal{F}_{x_i}^{corr}$ of $\mathcal{G}_s$ is computed by intra-graph convolution as follows:
\begin{equation}
   \mathcal{F}_{x_{i}}^{corr}=\sum_{j=1}^{N}\breve{\mathbf{E}}_{i, j} * f_{adj}(\mathcal{F}_{x_{j}})+f_{self}(\mathcal{F}_{x_{i}}),
\end{equation}
and likewise for $\mathcal{G}_t$. Here, $\breve{\mathbf{E}}$ is the L1 column-normalized adjacency matrix calculated from $\mathbf{E}$, and $f_{adj}$ and $f_{self}$ are message passing functions, which are implemented by fully connected layers and ReLU.

Second, the AIS module is used to calculate a soft correspondence matrix. The AIS module consists of an affinity layer, instance normalization and Sinkhorn. An affinity matrix $\mathbf{A}$ between the two graphs is computed as follows:
\begin{equation}
   \mathbf{A}_{i, j}=(\mathcal{F}_{x_{i}}^{corr})^{T} \mathbf{W}(\mathcal{F}_{y_{j}}^{corr}),
\end{equation}
where $\mathbf{W}$ is the learnable parameter in the affinity layer. If $\mathcal{F}_{x_i}^{corr}$,$\mathcal{F}_{y_j}^{corr}\in\mathbb{R}^{Q}$, then $\mathbf{W}\in\mathbb{R}^{Q\times Q}$. 

Before using Sinkhorn to compute the soft correspondence matrix $\widetilde{\mathbf{C}}$, we need to transform $\mathbf{A}$ into a normalized matrix while keeping the same distribution. There are two approaches to do so, and the naïve approach is to use softmax for rows or columns. The problem with this approach is that it processes each row or column and does not consider the matrix as a whole, which may result in the problem that a smaller value in $\mathbf{A}$ is transformed into a larger value in the transformed matrix \cite{ourcvpr}. To avoid this situation, we do not use softmax but use instance normalization \cite{ref44} to transform $\mathbf{A}$. Instance normalization not only considers all the elements globally, but also avoids the above problem caused by softmax. For handling outliers, we add an additional row and an additional column of ones to the transformed matrix and then input it into Sinkhorn \cite{ref36} to calculate the soft correspondence matrix $\widetilde{\mathbf{C}}$ by the iterative process of alternating row and column normalization.



Finally, we enhance the node features by exploring cross-correlation through cross-graph conv. Cross-graph conv is similar to intra-graph conv, except that features are aggregated from the node features of the other graph with edges replaced by $\widetilde{\mathbf{C}}$. The more similar the node pairs between the two graphs are, the higher the corresponding weight of $\widetilde{\mathbf{C}}$ will be. We obtain a new node feature $\mathcal{F}_{x_i}^\prime$ of node $x_{i}$ with a self-correlation feature and cross-correlation feature as follows:
\begin{equation}
   \mathcal{F}_{x_{i}}^{\prime}=f_{\text {cross}}(\mathcal{F}_{x_{i}}^{\text {corr}}, \sum_{j=1}^{\mathrm{M}} \widetilde{\mathbf{C}}_{i, j} * \mathcal{F}_{y_{j}}^{\text {corr}}),
\end{equation}
and likewise for $\mathcal{G}_t$. Here, $f_{cross}$ consists of a feature concatenate and a fully connected layer, and it is shared for $\mathcal{G}_s$ and $\mathcal{G}_t$. 

\subsection{Soft2Hard Correspondence Solver}\label{secLAP}
To compute the hard correspondence matrix ${\overline{\mathbf{C}}}^{pre}$, which is binary, we sum the elements of each row and each column of $\widetilde{\mathbf{C}}$ and take out the rows and columns with a sum greater than confidence threshold $\tau$, and apply a LAP solver based on Hungarian algorithm\cite{ref40} on the resulting matrix to obtain a binary matrix. Then, the elements of the binary matrix are assigned to a zero matrix with the shape of $\widetilde{\mathbf{C}}$ according to their position in $\widetilde{\mathbf{C}}$, and the result is the hard correspondence matrix ${\overline{\mathbf{C}}}^{pre}$ we need. Finally, for object-level registration, we take ${\overline{\mathbf{C}}}^{pre}$ as input to predict the transformation $\{\hat{\mathbf{R}},\hat{\mathbf{t}}\}$ by weighted SVD or RANSAC, and set $\tau=0.5$. Since scene-level registration is more difficult, we only use RANSAC as correspondence-based estimator to estimate the transformation parameters and set $\tau=0.7$. In addition, we also provided the influence of $\tau$ on scene-level registration in Section 6.3 of the supplementary materials.

\subsection{Loss}
Our loss function takes the ground truth correspondences directly as supervision, which is different from previous studies \cite{ref15,ref16,ref35} that define loss on transformation parameters. Focal loss \cite{lin2017focal} between soft correspondence matrix $\widetilde{\mathbf{C}}$ and ground-truth correspondence matrix $\overline{\mathbf{C}}^{gt}$ is adopted to train our model. The formula is as follows:

\begin{equation}
\begin{split}
   \text{loss}=-\sum_{i=1}^{N} \sum_{j=1}^{M} (\alpha (1- \widetilde{\mathbf{C}}_{i, j})^{\gamma} \overline{\mathbf{C}}_{i, j}^{gt} \log \widetilde{\mathbf{C}}_{i, j} \\ + (1 - \alpha)(\widetilde{\mathbf{C}}_{i, j})^{\gamma}(1-\overline{\mathbf{C}}_{i, j}^{gt})\log(1-\widetilde{\mathbf{C}}_{i, j})),
\end{split}
\end{equation}
where $\alpha$ is a balance factor to address class imbalance, $\alpha \in [0,1]$ for correct correspondences and $1 - \alpha$ for incorrect correspondences. $\gamma$ is a modulating factor to down-weight easy correspondences and thus focus training on hard ones to distinguish correspondences \cite{lin2017focal}. We empirically set $\alpha=0.5$ and $\gamma=0$ in object-level registration, which is equivalent to CE loss used in the conference paper \cite{ourcvpr}. Finding enough good correspondences for scene-level registration is more difficult, so we focus more on those hard correspondences and empirically set $\alpha=0.25$ and $\gamma=2$ as in \cite{lin2017focal}. Since our loss function is only related to the soft correspondence matrix $\widetilde{\mathbf{C}}$, the calculations in Section \ref{secLAP} do not need to be differentiable.

\subsection{Implementation Details}
This network is implemented using PyTorch and trained on a single Nvidia Tesla V100. We train the network using the SGD optimizer and set $L=2$ in this study. For object-level datasets, we consider a neighborhood of $K$ = 20 for local feature extractor and set $V=1024$ for high-dimensional local features. For scene-level datasets, we set $V=528$ for high-dimensional local features. The number of heads in multi-head attention for the Transformer in Edge Generator is 4. For all experiments, we only run one iteration during training. To achieve more precise registration, we iteratively update the transformation twice during test time. For more training details and network architecture implementation, please refer to the supplementary materials.

\section{Experiments}
Extensive experiments are conducted on four datasets, and this section is organized as follows. Firstly, we demonstrate the effectiveness of our registration method on object-level datasets ModelNet40 and ShapeNet with different settings. Next, we employ the scene-level datasets 3DMatch and 3DLoMatch to further evaluate the performance of our registration method in real scene-level registration. To show more real-world applications, experimental results on outdoor dataset KITTI are given in Section 4 of the supplementary materials. Finally, to demonstrate the importance of our two key components, we performed ablation studies on RGM and ModelNet40 datasets. The ablation studies include replacing distance matrix in AIS modules with L2 norm between node features, and using different methods to build edges instead of building edges by a transformer. 
\subsection{Object-level Registration}
\subsubsection{Benchmark Datasets}
\label{secDatasets_modelnet40}
\textbf{ModelNet40} The ModelNet40 \cite{ref45} dataset includes 12,311 meshed CAD models from 40 categories. We randomly sample 2,048 points from the mesh faces and rescale points into a unit sphere. Each category consists of official train/test splits. To select models for evaluation, we take 80$\%$ and 20$\%$ of the official train split as the training set and validation set, respectively, and the official test split for testing. For each object in the dataset, we randomly sample 1,024 points as the source point cloud $\mathbf{X}$, and then apply a random transformation on $\mathbf{X}$ to obtain the target point cloud $\mathbf{Y}$ and shuffle the point order. For the transformation applied, we randomly sample three Euler angles in the range of $[0,45]^{\circ}$ for rotation and three displacements in the range of $\left[-0.5,\ 0.5\right]$ along each axis for translation. Unless otherwise noted, these settings are used by default in all experiments.

\textbf{ShapeNet} The ShapeNet\cite{ref63} dataset collects 16,881 CAD shape models across 16 categories. The official splits are 14,006 models for training and 2,874 models for testing. The preprocessing of data is the same as ModelNet40, and testing of registration methods are only conducted on the test set.

\subsubsection{Evaluation Metrics}
For ModelNet40 dataset, we use six evaluation metrics, and the first four are calculated from the estimated transformation parameters. They are the mean isotropic errors (MIE) of $\mathbf{R}$ and $\mathbf{t}$ proposed in RPM-Net \cite{ref16}, and the mean absolute errors (MAE) of $\mathbf{R}$ and $\mathbf{t}$ used in DCP \cite{ref15}, which are anisotropic. All rotation-related metrics are in units of degrees.

In addition, we propose a new metric, clip chamfer distance (CCD), which measures how close the two point clouds are brought to each other, and it is calculated as follows:
\begin{align}
   \operatorname{CCD}(\widehat{\mathbf{X}}, \mathbf{Y}&)=\sum\limits_{\hat{x}_{i} \in \mathbf{X}}\min(\min\limits_{y_{j} \in \mathbf{Y}}(\left\|\hat{x}_{i}-y_{j}\right\|_{2}^{2}), d)
   \notag
   \\&\ \ +\sum\limits_{y_{j} \in \mathbf{Y}}\min(\min\limits_{\hat{x}_{i} \in \mathbf{X}}(\left\|\hat{x}_{i}-y_{j}\right\|_{2}^{2}), d),
\end{align}
where $\hat{\mathbf{X}}$ is the transformed source point cloud after registration and ${\hat{x}}_i$ is the $i$th point. To avoid the influence of outliers in partial-to-partial registration, the point pair whose distance is larger than 0.1 is not included in the calculation. This is implemented by setting the threshold $d=0.1$.

Finally, we also reported the recall with MAE($\mathbf{R}$) $<1^{\circ}$ and MAE($\mathbf{t}$) $< 0.1$. The best results are marked in bold font in tables.

\subsubsection{Comparing Methods}
We compare our method to ICP \cite{ref7}, fast global registration (FGR) \cite{ref25}, as well as three latest learning-based methods, RPM-Net \cite{ref16}, IDAM \cite{ref17} and DeepGMR \cite{ref35}. We also compare the methods proposed at the same time as our method, including OMNet \cite{xu2021omnet}, Predator \cite{ref58}. Other early learning-based methods, such as DCP and PointNetLK, are not directly compared, because experiments in \cite{ref16,ref17,ref35} have already shown that these new methods have better performance. We adopt the ICP and FGR implemented by Intel Open3D \cite{ref46}. For IDAM, DeepGMR, OMNet and Predator, we use the code provided by the authors and train the models according to the author's settings. For RPM-Net, we need to estimate the normals except in the clean experiment and use the code provided by the authors. The number of iterations of RPM-Net was set to 5 according to the author's article. ICP uses the identity matrix as initialization, and none of the other methods need transformation initialization. All models of the comparing learning-based methods are retrained because no trained model is available. We denote our method as RGM when using SVD as the Correspondence-based Estimator, and as RGM$^{\star}$ when using RANSAC as the Correspondence-based Estimator. In addition to using the default MLP network as the backbone, we also evaluate the performance of using KPFCN as the backbone, and the experimental results using KPFCN as the backbone are provided in the supplementary materials. 

\subsubsection{Clean Point Cloud}
We first evaluate the registration performance on clean point clouds and follow the sampling and transformation settings in Section \ref{secDatasets_modelnet40}. The ground-truth correspondences are obtained by the strict correspondences between $\mathbf{X}$ and $\mathbf{Y}$. All models are trained and evaluated on clean data, and Table~\ref{tab1} shows the performance of our method and its peers. Our method achieves the best performance and greatly outperforms the strongest learning-based method. In addition, the success rate of RGM reaches 100$\%$, and most of its error metrics are close to 0, which cannot be achieved by other existing methods. Although DeepGMR also achieves a 100$\%$ success rate, its errors are larger than RGM. Some qualitative comparisons are shown in Figure~\ref{fig4} (a).

\begin{table}[t]
   \caption{Performance on clean point clouds}\label{tab1}
   \resizebox{\columnwidth}{!}{
   \begin{tabular}{l|rrrrrr}
    \hline\hline
    Methods & MAE(R) & MAE(t) & MIE(R) & MIE(t) & CCD & Recall \\
    \hline
    ICP & $3.079$ & $0.02442$ & $6.4467$ & $0.05446$ & $0.03009$ & $74.19 \%$ \\
    FGR & $0.006$ & $0.00005$ & $0.0099$ & $0.00010$ & $0.00019$ & $99.96 \%$ \\
    RPM-Net & $0.109$ & $0.00050$ & $0.2464$ & $0.00112$ & $0.00089$ & $98.14 \%$ \\
    IDAM & $0.731$ & $0.01244$ & $1.3536$ & $0.02605$ & $0.04470$ & $75.81 \%$ \\
    DeepGMR & $0.001$ & $0.00001$ & $0.0156$ & $0.00002$ & $0.00003$ & $\mathbf{1 0 0 . 0 0 \%}$ \\
    OMNet & $0.471$ & $0.00468$ & $0.9511$ & $0.00985$ & $0.01318$ & $91.09 \%$ \\
    Predator & $0.302$ & $0.00393$ & $0.5716$ & $0.00833$ & $0.01572$ & $95.54 \%$ \\
    \hline 
    RGM & $<\mathbf{0 . 0 0 1}$ & $<\mathbf{0 . 0 0 0 0 1}$ & $\mathbf{0 . 0 0 9 6}$ & $<\mathbf{0 . 0 0 0 0 1}$ & $<\mathbf{0 . 0 0 0 0 1}$ & $\mathbf{1 0 0 . 0 0 \%}$ \\
    RGM$^{\star}$ & $<\mathbf{0 . 0 0 1}$ & $<\mathbf{0 . 0 0 0 0 1}$ & $0.0100$ & $<\mathbf{0 . 0 0 0 0 1}$ & $<\mathbf{0 . 0 0 0 0 1}$ & $\mathbf{1 0 0 . 0 0 \%}$ \\
    \hline
    \end{tabular}
    }
    $^{\star}$If there is a star, RANSAC is used as the estimator, otherwise SVD is used.
\end{table}

\begin{table}[t]
   \caption{Performance on point clouds with Gaussian noise}\label{tab2}
    \resizebox{\columnwidth}{!}{
   \begin{tabular}{l|rrrrrr}
    \hline\hline
    Methods & MAE(R) & MAE(t) & MIE(R) & MIE(t) & CCD & Recall \\
    \hline
    ICP & $3.127$ & $0.02256$ & $6.5030$ & $0.04944$ & $0.05387$ & $77.59 \%$ \\
    FGR & $5.405$ & $0.03386$ & $10.0079$ & $0.07080$ & $0.06918$ & $30.75 \%$ \\
    RPM-Net & $0.305$ & $0.00253$ & $0.5773$ & $0.00532$ & $0.04257$ & $96.68 \%$ \\
    IDAM & $1.818$ & $0.01416$ & $3.4916$ & $0.02915$ & $0.05436$ & $49.59 \%$ \\
    DeepGMR & $1.178$ & $0.00716$ & $2.2736$ & $0.01498$ & $0.05029$ & $56.52 \%$ \\
    OMNet & $0.939$ & $0.01511$ & $1.8557$ & $0.03090$ & $0.05169$ & $84.24 \%$ \\
    Predator & $0.515$ & $0.00481$ & $0.9633$ & $0.00961$ & $0.04469$ & $94.37 \%$ \\
    \hline
    RGM & $0.080$ & $0.00069$ & $0.1496$ & $0.00141$ & $0.04185$ & $99.51 \%$ \\
    RGM$^{\star}$ & $\mathbf{0 . 0 7 3}$ & $\mathbf{0 . 0 0 0 6 5}$ & $\mathbf{0 . 1 3 8 9}$ & $\mathbf{0 . 0 0 1 3 2}$ & $\mathbf{0 . 0 4 1 8 4}$ & $\mathbf{9 9 . 8 4 \%}$ \\
    \hline
    \end{tabular}
    }
\end{table}

\subsubsection{Gaussian Noise}
To evaluate the robustness to noise, Gaussian noise sampled from $\mathcal{N}\left(0,\ 0.01\right)$ and clipped to $\left[-0.05,\ 0.05\right]$ is independently added to each coordinate of the points in clean point clouds. These noises might destroy the original correspondences, so we need to rebuild them for training models that need ground truth correspondences. First, we compute the point pair distance between $\mathbf{Y}$ and $\mathbf{X}^\prime$, which is obtained by applying the ground truth transformation to $\mathbf{X}$. Then, if $x_i^\prime\in\mathbf{X}^\prime$ and $y_j\in\mathbf{Y}$ satisfy Eq.~\ref{eqmul}, they are regarded as a corresponding point pair and no longer appear in the next round calculation. Finally, we find corresponding point pairs again according to Eq.~\ref{eqmul} from the remaining points. To avoid long-distance point pairs being selected as a correspondence, we only consider the point pairs whose distance is less than 0.1. The reason why we find the corresponding point pair again from the remaining points is that the distance between the two points may not be the smallest but the second smallest, so they are not found in the first round.
\begin{equation}
   \min\limits_{x_{n}^{\prime} \in \mathbf{X}^{\prime}}(\left\|x_{n}^{\prime}-y_{j}\right\|_{2}^{2})=\left\|x_{i}^{\prime}-y_{j}\right\|_{2}^{2}=\min\limits_{y_{m} \in \mathbf{Y}}(\left\|x_{i}^{\prime}-y_{m}\right\|_{2}^{2}).\label{eqmul}
\end{equation}

\begin{figure*}[ht]
   \centering
   \includegraphics[width=1\linewidth]{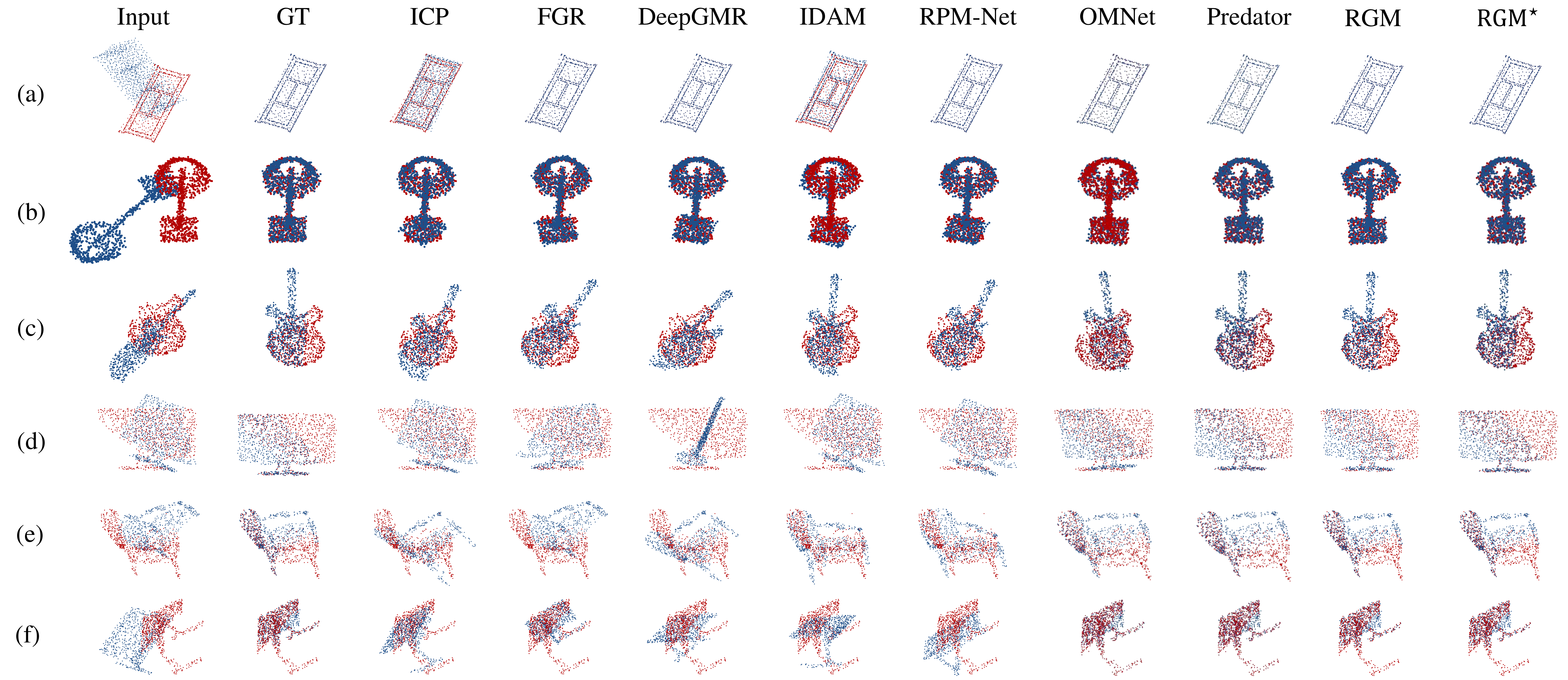}
   \caption{Qualitative registration results on ModelNet40, (a) clean, (b) noise, (c) partial-to-partial, and (d) unseen categories, (e) cross dataset, (f)full-range rotation.}
   \label{fig4}
\end{figure*}

All models are trained and evaluated on the noise data. The results are shown in Table~\ref{tab2}. It is obvious that our method is much more accurate than the latest learning-based methods and the traditional methods, and the recall of our method is close to 100$\%$. Some qualitative comparisons are shown in Figure~\ref{fig4} (b).

\begin{table}[t]
   \caption{Performance on partial-to-partial point clouds}\label{tab3}
   \resizebox{\columnwidth}{!}{
   \begin{tabular}{l|rrrrrr}
    \hline\hline
    Methods & MAE(R) & MAE(t) & MIE(R) & MIE(t) & CCD & Recall \\
    \hline
    ICP & $12.456$ & $0.12465$ & $24.8777$ & $0.26685$ & $0.11511$ & $6.56 \%$ \\
    FGR & $23.185$ & $0.14560$ & $42.4292$ & $0.30214$ & $0.12118$ & $5.23 \%$ \\
    RPM-Net & $0.864$ & $0.00834$ & $1.6985$ & $0.01763$ & $0.08457$ & $80.59 \%$ \\
    IDAM & $8.905$ & $0.09192$ & $16.9724$ & $0.19209$ & $0.12393$ & $0.81 \%$ \\
    DeepGMR & $43.683$ & $0.22479$ & $70.9143$ & $0.45705$ & $0.14401$ & $0.08 \%$ \\
    OMNet & $1.348$ & $0.01399$ & $2.6452$ & $0.02975$ & $0.08882$ & $71.55 \%$ \\
    Predator & $0.502$ & $0.00455$ & $0.9334$ & $0.00908$ & $0.08303$ & $93.47 \%$ \\
    \hline
    RGM & $0.492$ & $0.00414$ & $0.9298$ & $0.00874$ & $0.08238$ & $93.31 \%$ \\
    RGM$^{\star}$ & $\mathbf{0 . 3 0 8}$ & $\mathbf{0 . 0 0 2 7 5}$ & $\mathbf{0 . 6 1 0 9}$ & $\mathbf{0 . 0 0 5 7 4}$ & $\mathbf{0 . 0 8 2 2 1}$ & $\mathbf{9 5 . 5 4 \%}$ \\
    \hline
    \end{tabular}
    }
\end{table}

\subsubsection{Partial-to-Partial}

Partial-to-partial is the most challenging case for point cloud registration, and it is important because it occurs frequently in real-world applications.  To generate partial-to-partial point cloud pairs, we follow the protocol in RPM-Net \cite{ref16}, which is closer to real-world applications. For each point cloud, we create a random plane passing through the origin independently, translate it along its normal, and retain 70$\%$ of the points. All models are trained and evaluated on partial-to-partial data and the results are illustrated in Table~\ref{tab3}. Our method is obviously more accurate than the other methods, and its success rate is higher than 90$\%$. RPM-Net is the third best method, but its error is still twice as large as ours. Predator \cite{ref58} is significantly better than RPM-Net, but slightly worse than ours. Some qualitative comparisons are shown in Figure~\ref{fig4} (c). For the inference time of our method and the comparison methods, please refer to the supplementary materials.

\begin{table}[t]
   \vspace{-0.045cm}
   \caption{Performance on unseen categories point clouds}\label{tab4}
   \resizebox{\columnwidth}{!}{
   \begin{tabular}{l|rrrrrr}
    \hline\hline
    Methods & MAE(R) & MAE(t) & MIE(R) & MIE(t) & CCD & Recall \\
    \hline
    ICP & $13.326$ & $0.13033$ & $26.6447$ & $0.27774$ & $0.11879$ & $6.71 \%$ \\
    FGR & $23.950$ & $0.14067$ & $41.9631$ & $0.29106$ & $0.12370$ & $5.13 \%$ \\
    RPM-Net & $1.041$ & $0.01067$ & $1.9826$ & $0.02276$ & $0.08704$ & $75.59 \%$ \\
    IDAM & $10.158$ & $0.10063$ & $19.3249$ & $0.20729$ & $0.12921$ & $0.95 \%$ \\
    DeepGMR & $44.363$ & $0.22039$ & $71.0677$ & $0.44632$ & $0.14728$ & $0.24 \%$ \\
    OMNet & $1.742$ & $0.01766$ & $3.41012$ & $0.03772$ & $0.09291$ & $59.00 \%$ \\
    Predator & $0.517$ & $0.00499$ & $0.96325$ & $0.01014$ & $0.08431$ & $\mathbf{95.57\%}$ \\
    \hline
    RGM & $0.837$ & $0.00674$ & $1.5457$ & $0.01418$ & $0.08469$ & $84.28 \%$ \\
    RGM$^{\star}$ & $\mathbf{0 . 4 6 4}$ & $\mathbf{0 . 0 0 3 9 6}$ & $\mathbf{0 . 8 6 9 6}$ & $\mathbf{0 . 0 0 5 7 4}$ & $\mathbf{0 . 0 8 3 6 4}$ & $9 0 . 6 0 \%$ \\
    \hline
    \end{tabular}
    }
\end{table}

\begin{table}[t]
   \caption{Performance on Full-range Rotation point clouds}\label{tab5}
   \resizebox{\columnwidth}{!}{
   \begin{tabular}{l|rrrrrr}
    \hline\hline
    Methods & MAE(R) & MAE(t) & MIE(R) & MIE(t) & CCD & Recall \\
    \hline
    ICP & $79.5$ & $0.44$ & $138.2$ & $0.69$ & $0.138$ & $1.10 \%$ \\
    FGR & $72.6$ & $0.28$ & $101.5$ & $0.58$ & $0.143$ & $0.86 \%$ \\
    RPM-Net & $33.8$ & $0.09$ & $33.4$ & $0.21$ & $0.123$ & $4.19 \%$ \\
    IDAM & $73.1$ & $0.27$ & $83.2$ & $0.55$ & $0.165$ & $0.15 \%$ \\
    DeepGMR & $66.7$ & $0.21$ & $69.9$ & $0.45$ & $0.156$ & $<0.01 \%$ \\
    OMNet & $34.2$ & $0.11$ & $38.7$ & $0.23$ & $0.125$ & $17.85\%$ \\
    Predator & $3.1$ & $\mathbf{0.01}$ & $\mathbf{1.4}$ & $\mathbf{0.01}$ & $\mathbf{0.083}$ & $\mathbf{74.96 \%}$ \\
    \hline
    RGM & $19.1$ & $0.06$ & $15.9$ & $0.13$ & $0.101$ & $24.80 \%$ \\
    RGM$^{\star}$ & $\mathbf{2.9}$ & $\mathbf{0.01}$ & $1.5$ & $\mathbf{0.01}$ & $\mathbf{0.083}$ & $67.14 \%$ \\
    \hline
    \end{tabular}
    }
\end{table}

\subsubsection{Unseen Categories}
\label{sec:517}
To test each method’s generalization capability on unseen shape categories, we take the official train and test splits for the first 20 categories as the training and validation sets, respectively, and test on the official test splits of the last 20 categories. Other experimental settings are the same as those in the partial-to-partial experiment. The experimental results are summarized in Table~\ref{tab4}. We find that the performance of traditional methods does not change significantly. The generalization capability of RPM-Net is also good, but it is obvious that our method works better. Our method is superior to Predator \cite{ref58} in mean error but slightly lower in registration recall. The other learning-based methods do not generalize well to unseen categories. Some qualitative comparisons are shown in Figure~\ref{fig4} (d).

\subsubsection{Cross Dataset}
The purpose of this experiment is to validate the generalization ability of different methods between different object-level datasets. For every learning-based method, the model was trained on ModelNet40 while tested on ShapeNet. To more fully demonstrate the generalization of each model, we tested on three different settings: clean, Gaussian noise, and partial-to-partial. The test results for the different settings are summarized in Table \ref{tab15}. We can see that our method has no significant performance degradation on the ShapNet dataset, and the results of our method are much better than those of the other methods. Some qualitative comparisons are shown in Figure~\ref{fig4} (e).

\begin{table*}[t] \centering
   \caption{Performance of cross-dataset generalization (learning-based methods are trained on ModelNet40 and tested on ShapeNet)}\label{tab15}
   \resizebox{0.85\linewidth}{!}{
   \begin{tabular}{ l|l|rrrrrr}
   \hline\hline
   \multicolumn{1}{l}{Settings}      &Methods        & MIE(R) & MIE(t) & MAE(R) & MAE(t) & CCD & Recall \\
   \hline
   \multicolumn{1}{l}{} & ICP & $2.866$ & $0.02265$ & $5.7820$ & $0.04968$ & $0.01986$ & $85.56 \%$   \\
   \multicolumn{1}{l}{} & FGR & $0.001$ & $0.00001$ & $\mathbf{0.0010}$ & $0.00002$ & $0.00005$ & $\mathbf{100.00 \%}$ \\
   \multicolumn{1}{c}{} & RPM-Net & $0.139$ & $0.00093$ & $0.2847$ & $0.00198$ & $0.00186$ & $96.56 \%$ \\
   \multicolumn{1}{l}{} & IDAM & $7.756$ & $0.05623$ & $13.8014$ & $0.11481$ & $0.07644$ & $51.53 \%$ \\
   \multicolumn{1}{l}{Clean} & DeepGMR & $0.011$ & $0.00007$ & $0.0358$ & $0.00015$ & $0.00015$ & $99.79 \%$ \\
   \multicolumn{1}{l}{} & OMNet & $1.483$ & $0.07260$ & $3.0031$ & $0.14919$ & $0.07782$ & $53.54 \%$ \\
   \multicolumn{1}{l}{} & Predator & $0.406$ & $0.00483$ & $0.7677$ & $0.01014$ & $0.02011$ & $95.82 \%$ \\
   \cline{2-8}
   \multicolumn{1}{l}{} & RGM & $<\mathbf{0.001}$ & $<\mathbf{0.00001}$ & $0.0094$ & $<\mathbf{0.00001}$ & $<\mathbf{0.00001}$ & $\mathbf{100.00\%}$ \\
   \multicolumn{1}{l}{} & RGM$^{\star}$ & $<\mathbf{0.001}$ & $<\mathbf{0.00001}$ & $0.0104$ & $0.00001$ & $<\mathbf{0.00001}$ & $\mathbf{100.00\%}$ \\
   \hline
   \multicolumn{1}{l}{} & ICP & $2.893$ & $0.02319$ & $5.8934$ & $0.05097$ & $0.05173$ & $83.89 \%$ \\
   \multicolumn{1}{l}{} & FGR & $7.985$ & $0.05075$ & $13.6898$ & $0.10500$ & $0.08048$ & $29.23 \%$ \\
   \multicolumn{1}{l}{} & RPM-Net  & $0.429$ & $0.00363$ & $0.7936$ & $0.00775$ & $0.04111$ & $95.41 \%$ \\
   \multicolumn{1}{c}{} & IDAM & $1.818$ & $0.01565$ & $3.4658$ & $0.03177$ & $0.05380$ & $55.81 \%$ \\
   \multicolumn{1}{c}{ Gaussian } & DeepGMR & $1.065$ & $0.00653$ & $2.0160$ & $0.01380$ & $0.04885$ & $64.37 \%$ \\
   \multicolumn{1}{c}{ Noise } &OMNet & $0.996$ & $0.02268$ & $1.9571$ & $0.04725$ & $0.05312$ & $80.20 \%$ \\
   \multicolumn{1}{c}{} & Predator & $0.423$ & $0.00425$ & $0.7949$ & $0.00863$ & $0.04182$ & $95.12 \%$ \\
   \cline{2-8}
   \multicolumn{1}{l}{} & RGM & $0.143$ & $0.00122$ & $0.2810$ & $0.00257$ & $0.03995$ & $98.19 \%$ \\
   \multicolumn{1}{l}{} & RGM$^{\star}$ & $\mathbf{0.012}$ & $\mathbf{0.00097}$ & $\mathbf{0.2422}$ & $\mathbf{0.00204}$ & $\mathbf{0.03994}$ & $\mathbf{98.61\%}$ \\
   \hline
   \multicolumn{1}{l}{} & ICP & $12.207$ & $0.13438$ & $23.8867$ & $0.29026$ & $0.11853$ & $9.12 \%$ \\
   \multicolumn{1}{l}{} & FGR & $26.873$ & $0.18319$ & $47.4291$ & $0.37820$ & $0.13043$ & $5.08 \%$ \\
   \multicolumn{1}{c}{} & RPM-Net  & $2.917$ & $0.03698$ & $5.6309$ & $0.08199$ & $0.09150$ & $49.37 \%$ \\
   \multicolumn{1}{c}{Partial} & IDAM & $10.090$ & $0.10870$ & $19.2204$ & $0.22631$ & $0.13025$ & $1.81 \%$ \\
   \multicolumn{1}{c}{to} & DeepGMR & $46.901$ & $0.25167$ & $74.9521$ & $0.51363$ & $0.15012$ & $0.28 \%$ \\
   \multicolumn{1}{c}{Partial} & OMNet & $2.418$ & $0.04577$ & $4.7783$ & $0.09840$ & $0.10659$ & $36.81 \%$ \\
   \multicolumn{1}{c}{} & Predator & $0.589$ & $0.00644$ & $1.0931$ & $0.01377$ & $0.08162$ & $\mathbf{91.61 \%}$ \\
   \cline{2-8}
   \multicolumn{1}{l}{} & RGM & $0.887$ & $0.00791$ & $1.6236$ & $0.01696$ & $0.08201$ & $87.47 \%$ \\
   \multicolumn{1}{l}{} & RGM$^{\star}$ & $\mathbf{0.487}$ & $\mathbf{0.00446}$ & $\mathbf{0.9152}$ & $\mathbf{0.00929}$ & $\mathbf{0.08096}$ & $91.02\%$ \\
   \hline
   \end{tabular}
    }
\end{table*}

\subsubsection{Full-range Rotation}
Most existing learning-based object-level registration methods only consider the rotation within 45 degrees. However, in practice, the rotation between point cloud pairs can be any value within a full-range of [0,180]°. To compare the performance of each method within the [0,180]° rotation range, we conducted this full-range rotation experiment, in which we change the random sampling range of Euler angle to [0,180]°, and the other settings are consistent with partial-to-partial. We used the same rotation augmentations to train the model for all methods, and the results are summarized in Table \ref{tab5}. As the results show, all previous methods almost completely fail when the rotation range is [0,180]°. The previous best method, RPM-Net, has a recall of only 4.19$\%$. Our method can achieve a recall of 24.80$\%$ with SVD as the Correspondence-based Estimator, which is six times higher than RPM-Net. When we use RANSAC as the Correspondence-based Estimator, the recall can be further improved to 67.14$\%$. Compared with the most recent methods, OMNet \cite{xu2021omnet} and Predator \cite{ref58}, our method is significantly better than OMNet and comparable to Predator. Some qualitative comparisons under large rotation are shown in Figure~\ref{fig4} (f).

\begin{figure*}[t]
    \centering
    \includegraphics[width=1\linewidth]{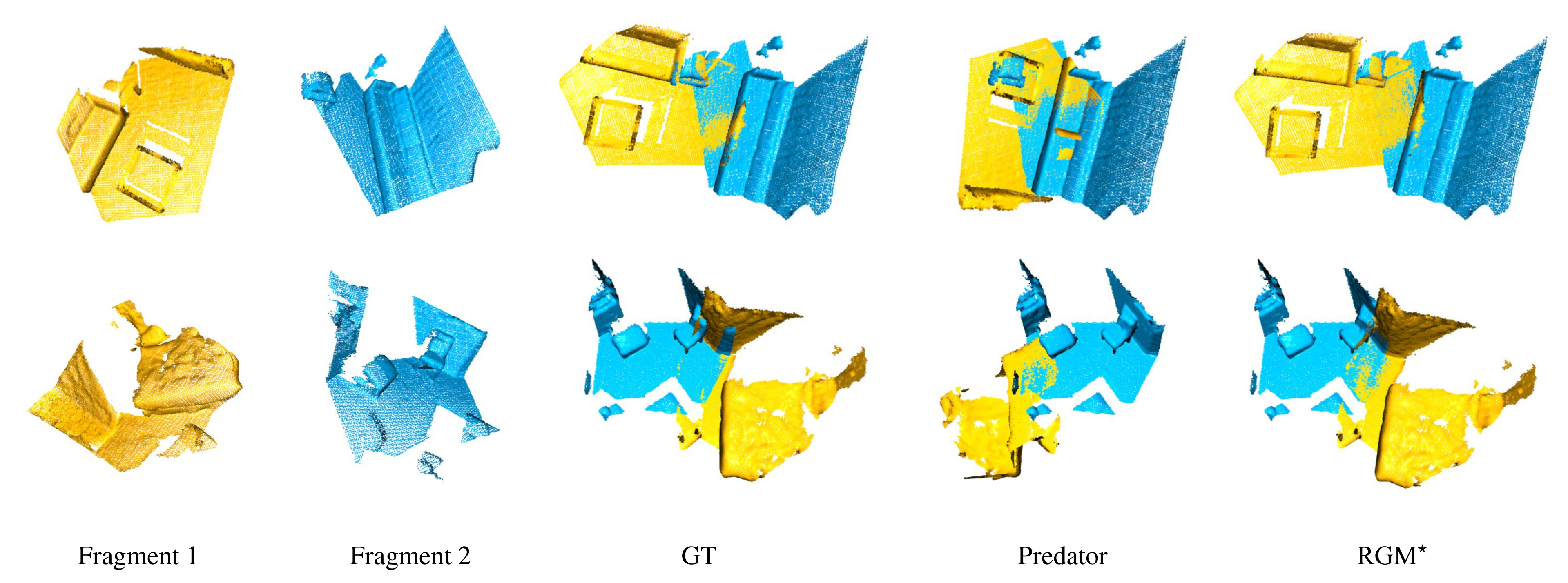}
    \caption{Qualitative registration results of RGM$^{\star}$ on 3DMatch.}\label{fig_vis_3dmatch}
\end{figure*}

\subsection{Scene-level Registration}
\subsubsection{Benchmark Datasets}
3DMatch \cite{ref55} is a benchmark of indoor rigid scan matching and registration, which includes 46 scenes for training, 8 scenes for validation and 8 for testing. 3DMatch considers only scan pairs with $>$30$\%$ overlap. However, in practice, there will be scan pairs with a lower overlap between 10 and 30$\%$. For this scenario, Huang \textit{et al.}\cite{ref58} constructed a more difficult dataset named 3DLoMatch.

\subsubsection{Evaluation Metrics}
According to the actual aim of point cloud registration, we use Registration Recall (RR), the percentage of successful alignment whose transformation error is smaller than a certain threshold (e.g., $\text{RMSE}<0.2 \mathrm{~m}$), as our main evaluation metric. Given a dataset $\mathcal{D}$ with $|\mathcal{D}|$ point cloud pairs, Registration Recall is calculated as follows:
\begin{equation}
\text{RR}(\mathcal{D})=\frac{1}{|\mathcal{D}|} \sum_{(\mathbf{X}, \mathbf{Y}) \in \mathcal{D}} \mathds{1}\left(\text{RMSE}(\mathbf{X}, \mathbf{Y})<\tau_{1}\right),
\end{equation}
where $\mathds{1}(\cdot)$ represents the indicator function, $\tau_{1}=0.2 \mathrm{~m}$ and for each scan pairs $(\mathbf{X}, \mathbf{Y}) \in \mathcal{D}$, $\text{RMSE}$ of the ground truth correspondence set $\overline{\mathbf{C}}^{gt}$ after applying the estimated transformation $\mathbf{T}^{pre}$ is defined as:
\begin{equation}
\text{RMSE}=\sqrt{\frac{1}{|\overline{\mathbf{C}}^{gt}|} \sum_{\left(x_{i}, y_{j}\right) \in \overline{\mathbf{C}}^{gt}}\left\|\mathbf{T}^{pre}\left(x_{i}\right)-y_{j}\right\|^{2}} .
\end{equation}
For keeping consistent with the original evaluation protocol of 3DMatch \cite{ref55}, immediately adjacent point clouds with very high overlap ratios are excluded. 
Following previous works \cite{ref60, ref61, ref58, ref62}, we also report two other metrics, namely Inlier Ratio (IR) and Feature Match Recall (FMR). The Inlier Ratio indicates the fraction of correspondences whose residual error in the geometry space is less than a threshold $\tau_{2}=10 \mathrm{~cm}$. Given the estimated correspondence set $\left(x_{i}, y_{j}\right) \in \overline{\mathbf{C}}^{pre}$, Inlier Ratio of a single scan pair is defined as:
\begin{equation}
\text{IR}=\frac{1}{|\overline{\mathbf{C}}^{pre}|} \sum_{\left(x_{i}, y_{j}\right) \in \overline{\mathbf{C}}^{pre}} \mathds{1}\left(\left\|{\mathbf{T}^{gt}}\left(x_{i}\right)-y_{j}\right\|_{2}<\tau_{2}\right),
\end{equation}
where ${\mathbf{T}^{gt}}$ indicates the ground truth transformation between $\mathbf{X}$ and $\mathbf{Y}$.
The Feature Match Recall is defined as the percentage of point cloud pairs whose Inlier Ratio is larger than $5\%$. 

\subsubsection{Comparing Methods}
We compare our method to  other state-of-the-art approaches in scene-level registration, including 3DSN \cite{ref59}, FCGF \cite{ref60}, D3Feat \cite{ref61}, Predator \cite{ref58}, SpinNet \cite{ao2021spinnet} and CoFiNet \cite{ref62} in Table \ref{3dmatch_table}. Other early methods, such as FPFH \cite{rusu2009fast}, PPFNet \cite{ref66} and RPM-Net \cite{ref16}, are not directly compared, because experiments in \cite{ref58, ref59} have already demonstrated that these new methods have better performance. The results of the comparing methods are the best results reported in the original paper. Because the scene-level dataset is complex and more challenging, the correspondence is not as good as that obtained in the object-level dataset, and the SVD performs poorly in scene-level. Therefore, like the comparison methods, our method only reports the results using RANSAC as the correspondence-based estimator.

\subsubsection{3DMatch and 3DLoMatch}
In this experiment, we evaluated the performance of our method in indoor scene and compared it with the state-of-the-art methods. We take KPFCN based $f_\theta$ described in Section \ref{sec:feature_extractor} as our feature extractor. Following previous literature \cite{ref59, ref58, ref62}, we apply a Hungarian algorithm\cite{ref40} based LAP solver on the corresponding matrix $\widetilde{\mathbf{C}}$ and find the transformation parameters with RANSAC for calculating RR. For a fair comparison with methods that randomly sample different points, we report the maximum Registration Recall(RR) for these methods. As presented in Table \ref{3dmatch_table}, there is no clear relationship between RR and FMR based on previous study \cite{ref58}. Although our method does not have the highest FMR and IR, it significantly outperforms existing methods in terms of RR, which is a more important metric in point cloud registration. This demonstrates that our method is more effective at reducing error correspondences, which can easily result in misregistration, than other methods. Considering the large variation of overlap rate in indoor scenes, we tried both the default CE loss and focal loss. Experiments show that our method with focal loss performs better in indoor scenes, outperforming the state-of-the-art method with a margin of 4.9$\%$ and 3.7$\%$ on 3DMatch and 3DLoMatch, respectively. Qualitative comparisons are shown in Figure \ref{fig_vis_3dmatch}.

\subsection{Visualization of Graph Edges}
\begin{figure}[t]
    \centering
    \includegraphics[width=1\linewidth]{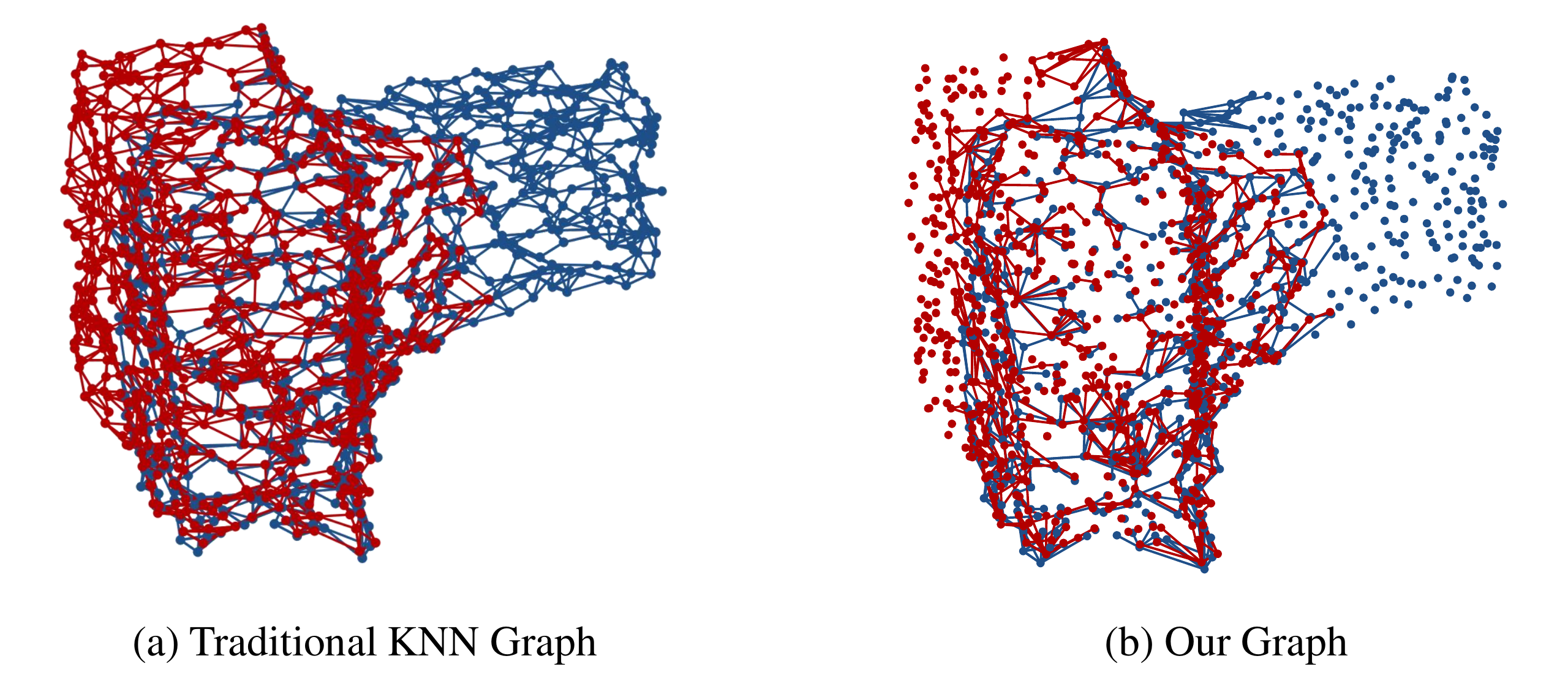}
    \caption{Visualization of graphs obtained by different edge generation methods. For our graph, down-sampled points and edges with large weights are illustrated. Please note that most edges of our graph are in the overlapping region and the edges of the two graphs roughly correspond to each other. These two properties are expected for better registration.}\label{fig10}
\end{figure}

In this section, we visualize the graph edges generated by different methods to demonstrate the superiority of building graph via our edge generator. The edges for an object model is shown in Figure~\ref{fig10}. For the KNN graph, we choose the five nearest neighbors of each point and draw an edge between it and the neighbors. For the proposed method, edges with a weight greater than 0.1 in the edge weight matrix are chosen for visualization. As illustrated in Figure~\ref{fig10}, our edge generator creates edges mostly in the overlapping parts, avoiding information interference in non-overlapping parts. Points in non-overlapping areas are not connected to any other points, making it easier to identify them.

\begin{table}[t]
   \caption{Performance on indoor datasets 3DMatch and 3DLoMatch}
   \label{3dmatch_table}
   \resizebox{\columnwidth}{!}{
   \begin{tabular}{l|ccc|ccc}
    \hline\hline
    & \multicolumn{3}{|c|}{ 3DMatch } & \multicolumn{3}{c}{ 3DLoMatch } \\
    & FMR & IR & RR & FMR & IR & RR \\
    \hline
    3DSN & $95.0$ & $36.0$ & $78.4$ & $63.6$ & $11.4$ & $33.0$ \\
    FCGF & $97.4$ & $56.8$ & $85.1$ & $75.4$ & $20.0$ & $41.7$ \\
    D3Feat & $95.4$ & $38.8$ & $84.5$ & $67.0$ & $14.0$ & $46.9$ \\
    Predator & $96.5$ & $57.1$ & $90.6$ & $76.3$ & $28.3$ & $62.4$ \\
    SpinNet & $97.6$ & $47.5$ & $88.6$ & $75.3$ & $20.5$ & $59.8$ \\
    CoFiNet & $\mathbf{98.1}$ & $49.8$ & $89.3$ & $\mathbf{83.1}$ & $24.4$ & $67.5$ \\
    \hline
    RGM$^{\star}$- CE & $96.7$ & $\mathbf{67.4}$ & $93.1$ & $79.6$ & $\mathbf{40.9}$ & $62.9$ \\
    RGM$^{\star}$- focal & $97.4$ & $50.5$ & $\mathbf{95.5}$ & $81.1$ & $22.4$ & $\mathbf{70.2}$ \\
    \hline
    \end{tabular}
    }
\end{table}

\subsection{Ablation Studies}

\begin{figure}[t]
   \centering
   \includegraphics[width=1\linewidth]{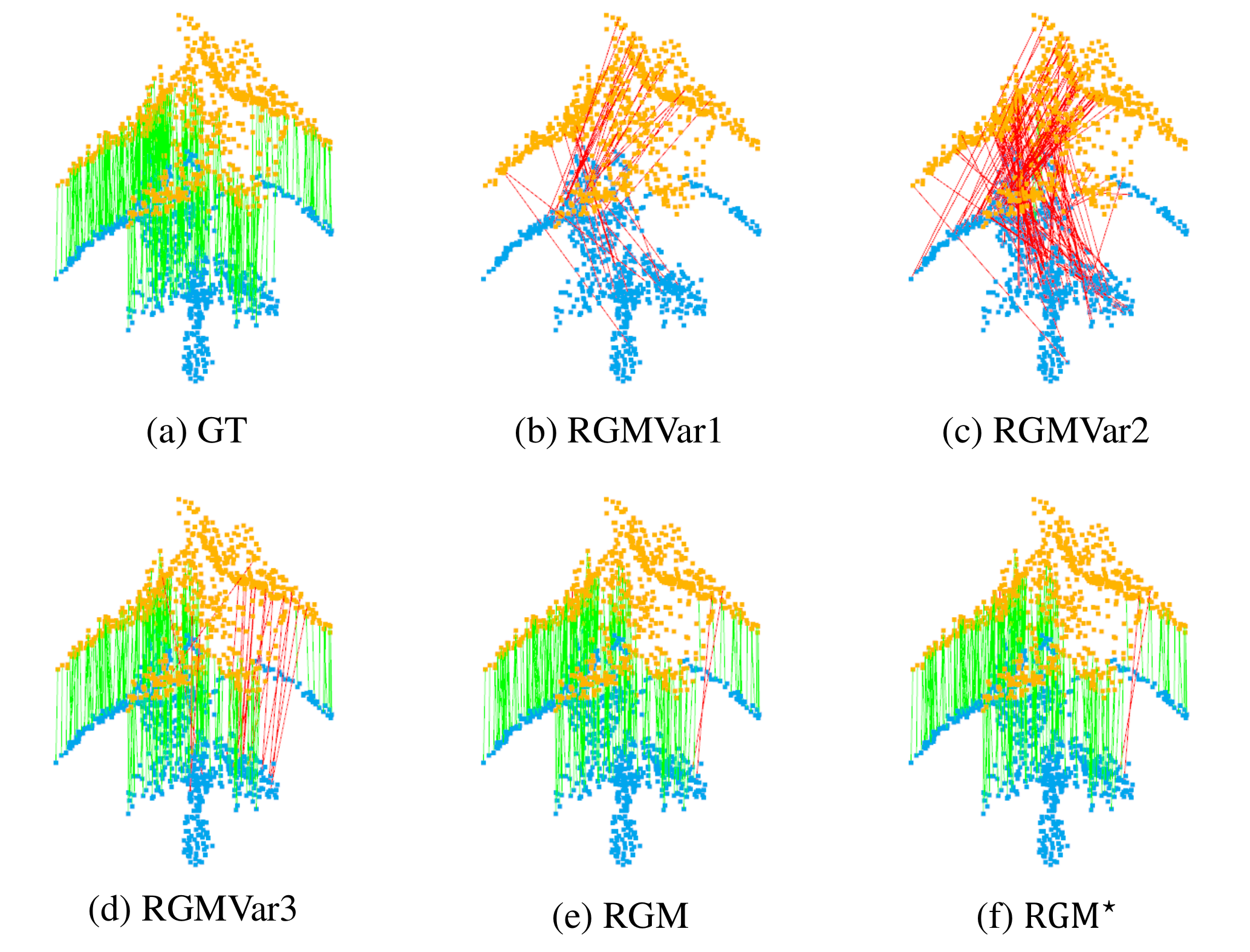}
   \caption{An illustrative case of the ground-truth correspondences and the correspondences generated by RGM and its variants. Much more correct correspondences are generated by RGM.}
   \label{fig5}
\end{figure}

In this section, we present the results of the ablation study to analyze the effectiveness of two key components. All ablation studies are performed on the partial-to-partial dataset with the same settings as section \ref{sec:517}. We analyze the two key components as follows:

To demonstrate the effectiveness of the AIS module, we design a variant to replace the AIS module, and the resulting method is denoted as RGMVar1. The variant computes the distance matrix $\mathbf{D}$ between the nodes of the two graphs by computing the L2 norm of node features, transforms $\mathbf{D}$ into a positive matrix within the finite values by the formula $e^{-\left(\mathbf{D}_{i,j}\ -\ 0.5\right)}$, and uses Sinkhorn to calculate the soft correspondences. The results are listed in the first row of Table~\ref{tab7}. We find that the registration accuracy becomes very poor by using the AIS variant, and this result shows that the proposed AIS module can effectively improve the registration performance. This is because the AIS module generates more correct matching than its variant, and an illustrative example of the correspondences generated by AIS and its variant is shown in Figure~\ref{fig5} (e) and (b).

\begin{table}[t]
   \caption{Ablation studies}\label{tab7}
   \resizebox{\columnwidth}{!}{
   \begin{tabular}{l|rrrrrr}
    \hline\hline
    Variants & MAE(R) & MAE(t) & MIE(R) & MIE(t) & CCD & Recall \\
    \hline
    RGMVar1 & $10.746$ & $0.07014$ & $19.2722$ & $0.14255$ & $0.11304$ & $18.33 \%$ \\
    RGMVar2 & $1.554$ & $0.01454$ & $2.9051$ & $0.03101$ & $0.08632$ & $74.17 \%$ \\
    RGMVar3 & $1.197$ & $0.01083$ & $2.2612$ & $0.02236$ & $0.08605$ & $75.59 \%$ \\
    RGM & $0.837$ & $0.00674$ & $1.5457$ & $0.01418$ & $0.08469$ & $84.28 \%$ \\
    RGM$^{\star}$ & $\mathbf{0 . 4 6 4}$ & $\mathbf{0 . 0 0 3 9 6}$ & $\mathbf{0 . 8 6 9 6}$ & $\mathbf{0 . 0 0 5 7 4}$ & $\mathbf{0 . 0 8 3 6 4}$ & $\mathbf{9 0 . 6 0 \%}$ \\
    \hline
    \end{tabular}
    }
\end{table}


To understand the importance of our edge generator, we design two variants that use full connection edges and sparse connection edges instead of building edges by a transformer, and the resulting methods are denoted as RGMVar2 and RGMVar3, respectively. For the full connection edges, we connect each point and all other points in the point cloud. For the sparse connection edges, we only connect each point and those points in a sphere centered at this point with a radius of 0.2. The results are shown in the second and third rows of Table~\ref{tab7}, and they are also inferior to the performance by using a transformer to generate edges. Examples of the hard correspondences generated by each of these two variants is shown in Figure~\ref{fig5} (c) and (d).




\section{Conclusion}

We introduce deep graph matching to solve the point cloud registration problem for the first time and propose a novel deep learning framework RGM that achieves state-of-the-art performance in both object-level and scene-level point cloud registration. We propose the AIS module to establish accurate correspondences between the graph nodes to greatly improve registration performance. In addition, the transformer-based edge generator provides a new idea for building graph edges in addition to full connection, nearest neighbor connection and Delaunay triangulation. We think that the deep graph matching approach has the potential to be used in other registration problems, including 2D-3D registration and deformable registration.


%


\ifCLASSOPTIONcompsoc
  \section*{Acknowledgments}
\else
  \section*{Acknowledgment}
\fi

This work was supported by the National Natural Science Foundation of China under Grant 62076070.


\bibliographystyle{IEEEtran}
\bibliography{egbib}

\end{document}